\documentclass{article}
\usepackage{acra}
\usepackage[table,dvipsnames]{xcolor}
\usepackage{graphicx}
\graphicspath{{Figs/}}
\usepackage{amsmath}
\usepackage{amssymb}
\usepackage{booktabs}
\usepackage{tikz}
\usetikzlibrary{arrows.meta, positioning, fit, backgrounds, calc}
\usepackage{soul}
\usepackage{cleveref}
\usepackage{url} 

\usetikzlibrary{arrows.meta, positioning, fit, backgrounds}
\usepackage{siunitx}
\usepackage{caption}
\newcommand\percentage[2][round-precision = 1]{%
  \SI[round-mode = places, scientific-notation = fixed, fixed-exponent = 0, #1]{#2e2}{\percent}%
}
\sethlcolor{yellow}

\definecolor{cTemporal}{HTML}{FCE5CD}
\definecolor{cLinguistic}{HTML}{D9EAD3}
\definecolor{cStructural}{HTML}{FFF2CC}
\definecolor{cGeometric}{HTML}{EAD1DC}

\DeclareRobustCommand{\hlc}[2]{\sethlcolor{#1}\hl{#2}}

\newif\ifSG
\SGtrue                                   %

\title{Beyond Appearance: Where Forest Visual Place Recognition Actually Fails}
\author{Blake Matheson${^{\star, 1}}$
        \quad \textbf{Ben McShanag$^{\star, 1}$}
        \quad \textbf{Isabella Moffat${^{\star, 1}}$} \\
        \textbf{Michael Milford$^2$}
        \quad \textbf{Peyman Moghadam$^{2,3}$}
        \quad \textbf{Dana Kuli\'c$^{1,3}$} \\
        \textbf{Somayeh Hussaini$^{\dagger, 2}$}
        \quad \textbf{Kavindie Katuwandeniya$^{\dagger, 1, 3}$}
        \\ [2pt] {\small $^{\star}$Equal contribution \quad $^{\dagger}$Equal supervision \quad
  $^{1}$Monash University \quad $^{2}$QUT \quad $^{3}$CSIRO Robotics}
  \thanks{Contact: \texttt{s.hussaini@qut.edu.au}, \texttt{kavindie.katuwandeniya@csiro.au}
}
}

\begin{document}

\maketitle

\begin{abstract}
Visual place recognition (VPR) in forest environments remains challenging because repetitive vegetation, and changes in illumination, season, and viewpoint can make distinct places appear similar and revisited places appear substantially different. In this paper, we investigate whether temporal, linguistic, structural, and geometric context can improve on an appearance-only descriptor, and what limits each representation when it does not. Using MegaLoc as the appearance-based baseline on the WildCross benchmark, we evaluate score-space and descriptor-space sequence matching, vision-language captions and reranking, scene-graph representations, and depth-derived stem graphs across same-direction revisits, extended routes, and reverse traversals. On a same-direction revisit, both sequence matching strategies reach similar Recall@1 (94.1\% vs.\ 91.5\%), but recall at 100\% precision rises from 50.9\% (single-frame MegaLoc) to 71.6\% in score space and falls to 36.5\% in descriptor space. The remaining context representations provide no consistent improvement over the appearance-based baseline. Same-place captions show limited place-specific discrimination, reranking is constrained by the initial candidate shortlist, and stem-graph matching remains weak even with LiDAR depth.

\end{abstract}

\section{Introduction}
\label{sec:intro}

Visual place recognition (VPR) is the task of identifying a previously visited place by matching a query image against a database of known places~\cite{lowry2015visual}. VPR is key to robot localisation and navigation, particularly in GNSS-denied environments, and enables loop closure for simultaneous localisation and mapping (SLAM), where robots must re-localise across changes in lighting, weather, and season~\cite{lowry2015visual}. 
Natural environments are particularly challenging for VPR due to repetitive vegetation, limited distinctive landmarks, and changes in illumination, season, and foliage, which can make different places appear similar (perceptual aliasing) and the same place appear different.
Benchmarks such as WildPlaces and WildCross highlight these challenges for LiDAR and visual place recognition~\cite{knights2023wild,knights2026wildcross}.

Despite advances in global descriptors, appearance alone often cannot distinguish similar places under changing conditions. Temporal context can reduce this dependence by exploiting consistency across consecutive observations, either in score space or descriptor space~\cite{milford2012seqslam,garg2021seqnet}. Other approaches use language, semantic structure, or geometry to add representations beyond a single global descriptor, but these have mainly been studied in urban, indoor, or landmark-rich environments. Their effectiveness in repetitive natural scenes remains less clear. 
Since a single false loop closure can corrupt a map~\cite{lowry2015visual}, we report both Recall@1 (R@1), the fraction of queries whose top match is correct, and recall at 100\% precision (R@100P), the fraction of queries accepted at the strictest threshold with no false positives.

In this work, we investigate where these complementary sources of information improve VPR in forests and, importantly, what limits them when they do not. We organise the analysis around four forms of information: temporal context, which captures how observations evolve along a trajectory; linguistic context, which describes scene content; structural context, which represents objects and their image-plane arrangement; and geometric context, which captures metric scene layout. 
Beyond retrieval performance, we analyse which failure modes each representation addresses and what limits its effectiveness in forests.
Our key contributions are:

\begin{itemize}
    \item We compare two training-free, temporal  sequence-matching strategies for forest VPR, applying temporal context in either score space or descriptor space. 
    \item We evaluate linguistic, geometric, and structural representations as complements to appearance-based VPR. We identify the main factors limiting their effectiveness in forests, including weak place discrimination in captions, candidate-recall limits in reranking, limited discriminative value in 2D scenes, and geometric reconstruction error in stem graphs.
    \item We evaluate these approaches on WildCross across long-term revisits, extended routes, and reverse traversals, and analyse which forms of context remain effective under each condition.

\end{itemize}

\section{Related Work}
\label{sec:related}

\subsection{Global Descriptors for VPR}
\label{subsec:rel-descriptors}
VPR systems encode each image into a compact descriptor and retrieve the database image whose descriptor is nearest~\cite{lowry2015visual}, so performance hinges on descriptor robustness to appearance and viewpoint changes. 
CNN-based methods established the template, including NetVLAD's learnable VLAD aggregation~\cite{arandjelovic2016netvlad} and CosPlace's classification reframing for large-scale training~\cite{berton2022rethinking}. However, being trained largely on urban imagery, they often generalise poorly to unstructured environments~\cite{knights2026wildcross}.
Foundation models such as DINOv2~\cite{oquab2024dinov} provide more general-purpose features, and current methods primarily differ in how they aggregate and adapt them. AnyLoc~\cite{keetha2023anyloc} pairs frozen DINOv2 with VLAD and targets unstructured environments, 
and MegaLoc~\cite{berton2025megaloc} trains on a unified mixture of VPR, landmark, and localisation data.
Nonetheless, place recognition in forest environments remains challenging~\cite{keetha2023anyloc}, which motivates our work.

\subsection{Sequence Matching}
\label{subsec:rel-seq}
Sequence matching exploits trajectory order in three main ways: searching the similarity matrix for an aligned trajectory, aggregating single-image descriptors into a sequence descriptor, or learning temporal context during training. 
SeqSLAM~\cite{milford2012seqslam} searches the similarity matrix for the best-scoring path over consecutive frames, assuming a roughly constant velocity ratio and travel direction, so degrades under velocity mismatch or reverse traversal. 
SeqNet~\cite{garg2021seqnet} aggregates consecutive descriptors into a sequence descriptor to retrieve candidate sequences before matching frames within them.
JIST~\cite{berton2023jist} uses joint training on single images and sequences, but on urban data.
As large forest data are scarce, evaluating learned models on familiar environments risks rewarding memorisation over generalisation.
Thus, we use training-free methods, i.e., SeqSLAM-style score-space search and SeqNet-style descriptor averaging without learned layers.

\subsection{Reranking and Failure Prediction}
\label{subsec:rel-rerank}
A standard approach beyond a single global descriptor is a two-stage pipeline in which retrieval proposes a shortlist of candidates and a second stage reorders it, using, for example, a learned pair classifier as in Pair-VPR~\cite{hausler2025pair}, or training-free local matching with SuperPoint~\cite{detone2018superpoint} and LightGlue~\cite{lindenberger2023lightglue}. 
Any such pipeline is bounded by the performance of its first stage, which we measure explicitly.
An important consideration is determining \emph{which} queries should be selected for reranking, since it is especially beneficial when the initial retrieval is incorrect and can otherwise degrade correct predictions. Training-free uncertainty estimates have been proposed based on retrieval scores~\cite{miller2026through} or the spatial dispersion of shortlisted reference poses~\cite{zaffar2024estimation}, although the latter assumes access to geotagged references. We instead use a gating strategy that combines score-based and temporal cues, and does not require reference poses.

\begin{figure*}[t]
    \centering
    \includegraphics[width=0.9\textwidth]{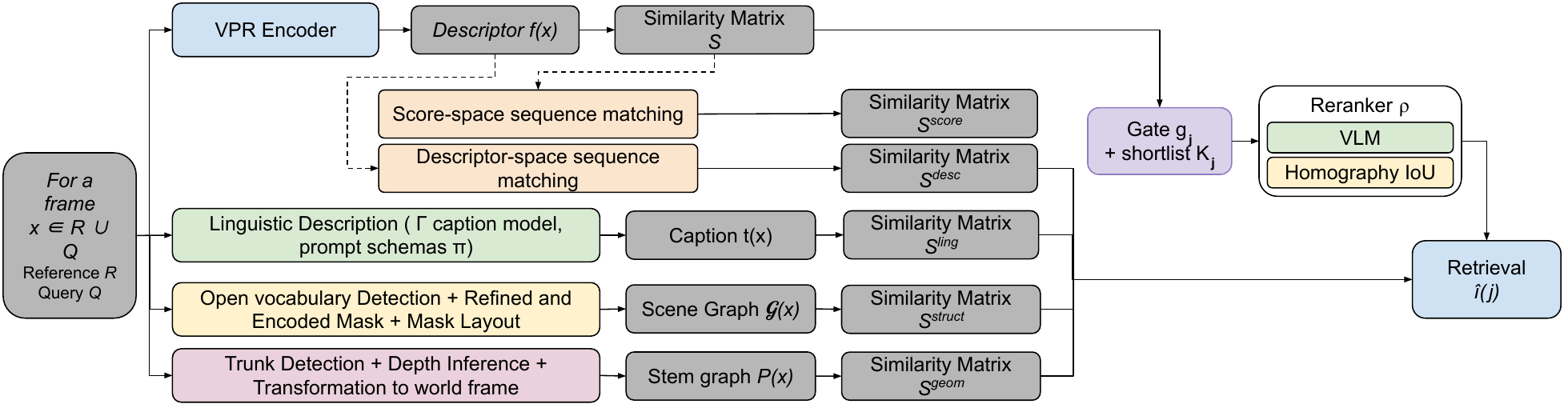}
    \caption{
    The four diagnostic axes, each defined by the space in which similarity is measured. The \hlc{cTemporal}{temporal axis} acts on the appearance pipeline itself, aggregating either the descriptors or the similarity matrix; the \hlc{cLinguistic}{linguistic}, \hlc{cStructural}{structural}, and \hlc{cGeometric}{geometric} axes branch from the frames and produce their own similarity matrices. A gate on the similarity matrix $S$ selects queries whose shortlist is reordered by a reranker $\rho$ on images. Each output is compared against the baseline $S$ on identical traversals.
}
    \label{fig:overview}
    \vspace{-0.4cm}
\end{figure*}

\subsection{Language for Place Recognition}
\label{subsec:rel-vlm}

Vision-Language Models (VLMs) have been used in VPR as both descriptors and rerankers. As a \textit{descriptor}, language is fused with visual embeddings; LaVPR~\cite{idan2026lavpr} reports improved robustness under visually degraded conditions. As a \textit{reranker}, VLMs reorder candidates from an initial visual retrieval stage. LLM4VPR~\cite{lyu2024tell} reasons over textual descriptions of query--candidate differences, while Scale-Don't-Fine-tune~\cite{cheng2025scale} scores image pairs directly to avoid losing discriminative visual cues through text generation. We use captions to assess how much place-specific information is preserved in language, while the reranker operates directly on images. Since prior work primarily targets urban environments with distinctive semantic cues, we evaluate them zero-shot in visually repetitive forest scenes. 

\ifSG
\subsection{Structure for Place Recognition}
\label{subsec:rel-sg}

Scene graphs represent places through objects and their spatial relations, providing structural information beyond global image descriptors. Prior work has matched scene graphs to OpenStreetMap~\cite{jung2025gotpr}, learned joint place-object representations from unposed RGB~\cite{zhang2024multiview}, and used open-set segments to improve viewpoint robustness~\cite{garg2024revisit}, primarily in urban, driving, or indoor environments. In forestry robotics, tree trunks are commonly localised from visual data for mapping and navigation~\cite{pierzchala2018mapping}, but are rarely used as place descriptors. \Cref{subsec:stem-method} explores this direction for forest place recognition. 

\fi

\section{Methodology}
\label{sec:method}

\subsection{Overview and Problem Formulation}
\label{subsec:overview}

We compare frames along axes defined by the information used to measure similarity (\Cref{fig:overview}): \emph{temporal}, over short sequences along the trajectory (\Cref{subsec:seq-method}), and \emph{linguistic}, between VLM captions (\Cref{subsec:vlm-method})\ifSG{}; \emph{structural}, over detected objects and their image-plane arrangement (\Cref{subsec:sg-method}); and \emph{geometric}, over tree-trunk positions in metric ground-plane coordinates (\Cref{subsec:stem-method})\fi{}. The temporal stream directly replaces the appearance baseline. The remaining streams are evaluated as diagnostics, and for each we report what limited it in our implementation.

Let $\mathcal{R} = \{r_i\}_{i=1}^{N}$ be an ordered reference traversal and $\mathcal{Q} = \{q_j\}_{j=1}^{M}$ be an ordered query traversal of the same environment. 
Each frame $x$ is encoded by an off-the-shelf VPR model $F$ and L2-normalised, $\mathbf{d} = F(x)/\lVert F(x)\rVert_2 \in \mathbb{R}^{D}$; we write $\mathbf{r}_i$ and $\mathbf{q}_j$ for the descriptors of $r_i$ and $q_j$. Stacking them as columns of $\mathbf{X}_R \in \mathbb{R}^{D \times N}$ and $\mathbf{X}_Q \in \mathbb{R}^{D \times M}$ gives the cosine similarity matrix:
\begin{equation}
S = \mathbf{X}_R^{\top} \mathbf{X}_Q \in \mathbb{R}^{N \times M}, \qquad S_{ij} = \mathbf{r}_i^{\top}\mathbf{q}_j.
\label{eq:similarity}
\end{equation}
Single-frame retrieval, our baseline throughout, returns $\hat{\imath}(j) = \arg\max_i S_{ij}$.
Each diagnostic axis produces its own similarity matrix $S^{(a)} \in \mathbb{R}^{N \times M}$, evaluated against $S$ on the same traversals. The linguistic stream is also used as a rerank operator $\rho$, which reorders a shortlist of candidates only for queries flagged by a gate $g_j \in \{0,1\}$ computed from $S$. This limits how many already-correct queries a weaker representation can disturb.

\subsection{Temporal Stream: Sequence Matching} %
\label{subsec:seq-method}

Temporal context can be added at either stage of a monocular VPR pipeline: to the similarities, following SeqSLAM~\cite{milford2012seqslam} (\emph{score space}), or to the descriptors before $S$ is formed, similar to SeqNet~\cite{garg2021seqnet} (\emph{descriptor space}). We study how the choice of stage affects R@1 and R@100P.

\subsubsection{Score-Space Matching} %
Score-space matching adds temporal context after the similarity matrix is formed, without changing the descriptors.
Although a single forest frame can resemble many places, if $r_i$ is the correct match for $q_j$, the query frames leading up to $q_j$ should also match the reference frames around $r_i$.
Thus, a correct match lies on a trajectory of consistently high similarity in $S$, whereas a false match caused by visual self-similarity tends to be isolated.
Following SeqSLAM~\cite{milford2012seqslam}, we re-score each cell $(i,j)$ by how well it is supported by such a trajectory.
The trajectory covers the $L$ query frames ending at $q_j$. For a slope $v$, it visits reference frame $i - \lfloor vk \rceil$ at query frame $j - k$, for $k = 0, \dots, L-1$, where $\lfloor\cdot\rceil$ rounds to the nearest integer.
The slope $v$ is the number of reference frames the trajectory advances per query frame. A slope of $v = 1$ means both traversals have the same frame spacing. Other positive values compensate for a spacing mismatch, and negative values indicate a reverse traversal, where stepping back through the query steps forward through the reference.
We average $S$ along the trajectory and keep the best slope from a candidate set $\mathcal{V}$,
\begin{equation}
S^{\text{score}}_{ij} = \max_{v \in \mathcal{V}} \frac{1}{L} \sum_{k=0}^{L-1} S_{i - \lfloor vk \rceil,j - k}.
\vspace{-0.2cm}
\label{eq:seqslam}
\end{equation}
Near the edges of $S$, only the cells that fall inside the matrix are averaged, so the first $L-1$ query frames are scored with a shorter trajectory.
For each reference-query pairing, $\mathcal{V}$ holds evenly spaced slopes that bracket the ratio of the median query frame spacing to the median reference frame spacing, so that the true relative speed lies within the search.
We add the negatives of these slopes to cover reverse traversal without the direction having to be known in advance. 
We write $S^{\text{score}} = \mathcal{T}_{\text{score}}(S)$, and retrieve with $\hat{\imath}(j) = \arg\max_i S^{\text{score}}_{ij}$. 

\subsubsection{Descriptor-Space Matching} %
Here we add the temporal context to the descriptors. Each L2-normalised reference descriptor is replaced by the re-normalised mean of the descriptors in a trailing window of $h$ frames ending at it,
\begin{equation}
\bar{\mathbf{r}}_i = \frac{\sum_{i' \in \mathcal{W}_i} \mathbf{r}_{i'}}{\big\lVert \sum_{i' \in \mathcal{W}_i} \mathbf{r}_{i'} \big\rVert_2},  \mathcal{W}_i = \{ i' : i - h < i' \le i \} \cap \{1, \dots, N\},
\label{eq:descagg} %
\end{equation} 
and each query descriptor $\bar{\mathbf{q}}_j$ is obtained in the same way over $\mathcal{Q}$. Writing $\bar{\mathbf{X}} = \mathcal{T}_{\text{desc}}(\mathbf{X})$ for this aggregation applied to a descriptor matrix $\mathbf{X}$, we form the similarity matrix from the aggregated descriptors,
\begin{equation}
S^{\text{desc}} = \mathcal{T}_{\text{desc}}(\mathbf{X}_R)^{\top} \mathcal{T}_{\text{desc}}(\mathbf{X}_Q) = \bar{\mathbf{X}}_R^{\top} \bar{\mathbf{X}}_Q,
\label{eq:descsim}
\end{equation}
and retrieve with $\hat{\imath}(j) = \arg\max_i S^{\text{desc}}_{ij}$. In contrast to score-space matching, this aggregation has no slope parameter. It implicitly assumes a near-unity slope and forward traversal, so it cannot adapt when the two traversals advance at different rates.

\ifSG
  \newcommand{\sgcapclause}{The scene-graph stream contributes an independent similarity matrix.}
\else
  \newcommand{\sgcapclause}{}
\fi

\subsection{Complementary Representations}
\label{subsec:complementary}

Forest VPR fails mainly through perceptual aliasing and viewpoint change, such as reverse traversal~\cite{knights2026wildcross}, so we test whether representations that capture different information can recover these failures.

\subsubsection{Linguistic Stream: Captions and Reranking}
\label{subsec:vlm-method}

We test a VLM in two roles: as a caption-based replacement for the image descriptor, and as a reranker that compares the query image directly with the baseline's top candidates.

\paragraph{Caption Retrieval.}
Each frame $x$ is captioned once, $t(x) = \Gamma(x, \pi)$, for a caption model $\Gamma$ and prompt schema $\pi$. Then, frames are compared by a text similarity function $\phi$,
\begin{equation}
S^{\text{ling}}_{ij} = \phi ( t(r_i), t(q_j)).
\label{eq:ling-sim}
\end{equation}

We sweep three axes: $\Gamma$, $\pi$ and $\phi$. The first is the caption model ($\Gamma$), comparing a locally run Qwen2.5-VL 7B~\cite{bai2025qwen25vltechnicalreport} against GPT-5.6 Sol~\cite{openai2026gpt56}. The second is the prompt ($\pi$), using three schemas ranging from a 15-field tag list, through a six-part structured description, to an eight-section essay.
The third is the scoring function ($\phi$), using eight retrieval scorers covering semantic embeddings (BGE-M3 in its dense, sparse and multi-vector modes~\cite{chen2024bge}), lexical overlap (TF-IDF~\cite{salton1988term} and BM25~\cite{robertson2009probabilistic}), and captioning metrics (BLEU-4, METEOR, ROUGE-L)~\cite{chen2015microsoft}, so that a failure common to all eight cannot be attributed to the scorer. Recall@1 is computed tie-aware, since schema-constrained captions can produce byte-identical text for distinct places.

\paragraph{VLM Reranking.}
Here, the VLM reorders the baseline's shortlist rather than replacing it. 
The \emph{candidate picker} selects a shortlist $\mathcal{K}_j$ from the baseline ranking of $S_{\cdot j}$.
Unless stated otherwise, the picker is a tie band.
$\mathcal{K}_j$ holds the reference frames whose cosine similarity to $q_j$ is within 0.0252 of the top-1 score, up to the ten highest-scoring.
One configuration uses a fixed top-10 shortlist. 
On gated queries with at least two candidates, the VLM returns a permutation $\rho(\mathcal{K}_j)$, written back over those entries of $S_{\cdot j}$, so:
\begin{equation}
\hat{\imath}(j) =
\begin{cases}
\rho ( \mathcal{K}_j)_1, & g_j = 1 \text{ and } |\mathcal{K}_j| \ge 2,\\[2pt]
\arg\max_i S_{ij}, & \text{otherwise.}
\end{cases}
\vspace{-0.2cm}
\label{eq:ling-rerank}
\end{equation}

Thus, reranking changes only gated queries. 
To probe the effect of model capacity, we smoke-test three open-weight 8-12B models on five randomly selected queries, scale the strongest, Qwen3-VL \cite{bai2025qwen3}, to 32B, and compare both against GPT-5.6 Sol \cite{openai2026gpt56} on identical shortlists. Listwise ranking is order-sensitive~\cite{liu2024lost}, as the smoke tests confirmed, so all full runs present candidates in a single fixed order, sorted by baseline similarity. 

\paragraph{Reranking Failure Gate.}
\label{subsec:gate-method} 
The failure gate is a training-free, pose-free detector that rank-averages the spread of the top-10 similarity scores with an along-track residual computed in frame-index space, so it needs only the similarity matrix and database frame order. It sets $g_j = 1$ for a fixed budget of the highest-ranked queries rather than thresholding a score. 

\ifSG

\subsubsection{Structural Stream: Scene Graphs}
\label{subsec:structural_geometric}

This stream tests whether describing a frame by its objects, rather than by one global embedding, separates places that the embedding confuses. 
Arrangement is measured only in the image plane, where camera motion and scene change can look alike; the geometric stream below instead compares trunk positions in metric ground-plane coordinates.

\paragraph{Scene Graph Descriptors.} %
\label{subsec:sg-method}

Graphs are built zero-shot, with no forest-specific training. For a frame $x \in \mathcal{R} \cup \mathcal{Q}$, an open-vocabulary detector (Grounding-DINO~\cite{liu2024grounding}) is prompted with an 18-class forest vocabulary $\mathcal{C}$ (trunk, canopy, trail, fallen log, understory, sky, and others) and the $K = 12$ most confident detections are retained. Each is refined into a mask with SAM2~\cite{ravi2025sam}, and its masked crop is encoded by DINOv2~\cite{oquab2024dinov}, giving:
\begin{equation}
\mathcal{G}(x) = ( \mathcal{O}_x, \mathcal{E}_x ), \qquad
\mathcal{O}_x = \{ (c_n,\, \mathbf{b}_n,\, \boldsymbol{\psi}_n) \}_{n=1}^{|\mathcal{O}_x|},
\vspace{-0.2cm}
\label{eq:sg-graph}
\end{equation}
where $c_n \in \mathcal{C}$, $\mathbf{b}_n$ and $\boldsymbol{\psi}_n \in \mathbb{R}^{768}$ are the class, box, and feature of object $n$. Each pair of objects of different classes adds one edge to $\mathcal{E}_x$, labelled by containment, proximity, or the dominant direction of separation. Such coarse labels are more likely than metric offsets to survive a change of viewpoint.

We read three fixed-length descriptors from each graph. The \emph{mean-pool} descriptor averages all object features, keeping the appearance of detected objects only. The \emph{class-binned} descriptor gives each class a slot holding the mean feature of its instances, with zero where the class is absent, so trunk appearance is compared against trunk appearance rather than blended with the canopy. The \emph{layout} descriptor discards visual features entirely, concatenating class counts, box position and area statistics, relation counts, and the object and edge totals. Each is L2-normalised and yields a cosine similarity matrix; the three are averaged to give $S^{\text{struct}}$.

\paragraph{Segment Reranking.}
We also test image-plane correspondence as a rerank operator $\rho$ over the ten best baseline candidates. For each query-candidate pair, SuperPoint~\cite{detone2018superpoint} keypoints are matched with LightGlue~\cite{lindenberger2023lightglue}, a homography is fitted with MAGSAC~\cite{barath2020magsac++}, the query's SAM2 masks are warped into the candidate view, and candidates are reordered by mask intersection over union (IoU). A homography holds only for a planar scene or a purely rotating camera, so we expect this signal to weaken as the distance between viewpoints grows.

\subsubsection{Geometric Stream: Stem Graphs} %
\label{subsec:stem-method}

This stream keeps only where tree trunks stand on the ground and discards appearance (\Cref{fig:stemgraph}). Trunks are vertical, persistent and sparse, which makes them good landmarks. This extends trunk localisation from forestry robotics~\cite{pierzchala2018mapping} to place recognition. 
\paragraph{Stem-Graph Construction.}
Each frame is reduced independently, without global pose, to the layout of trunks around the camera.
\emph{Input:} one RGB image.
\emph{Output:} the stem graph $\mathcal{P}(x) = \{(X_k, Z_k, d_k)\}_{k=1}^{K}$ of $K \approx 30$ trunks, each with a ground-plane position $(X_k, Z_k)$ in metres and a diameter $d_k$, in the camera frame.
We prompt SAM3~\cite{carion2026sam} with \emph{tree trunk} for trunk masks, and estimate metric depth with Depth Anything 3~\cite{lin2026depth} over an eight-frame window so several views constrain each depth map. From the lowest quarter of each mask, where the trunk meets the ground, we read the median depth, centre column and pixel width, and back-project them through the predicted intrinsics to a position and diameter. This assumes a level camera, so roll and pitch on uneven ground distort the layout. Trunks beyond 25m are discarded.

\paragraph{Stem Graph Matching.}
This stage scores whether two stem graphs show the same place.
\emph{Input:} a reference and a query stem graph, $\mathcal{P}(x_i)$ and $\mathcal{P}(x_j)$. \emph{Output:} a scalar similarity $S^{\text{geom}}_{ij}$.
We perform this matching in two ways. 
First, iterative closest point (ICP)~\cite{besl1992method} matches individual trunks. It alternates assigning each query trunk its nearest reference trunk with a closed-form solve for rotation, uniform scale and translation over the closest $80\%$ of pairs, so trunks seen in only one frame do not skew the alignment. As the relative heading is unknown, we keep the lowest residual over $16$ evenly spaced starting headings, and $S^{\text{icp}}_{ij}$ is the negated mean distance between aligned trunks.
Second, the \emph{bird's-eye-view} (BEV) descriptor summarises each graph. Trunks are rasterised onto a ground-plane grid of 0.25m cells, each adding its diameter to its cell. The grid is blurred with a Gaussian ($\sigma = 1.0$m) so that a trunk displaced by depth error still overlaps its own mark, then flattened and L2-normalised, and cosine similarity gives $S^{\text{bev}}$. BEV needs only one matrix product but, unlike ICP, is sensitive to camera rotation. The geometric stream is thus $S^{\text{geom}} \in \{ S^{\text{icp}}, S^{\text{bev}} \}$.

\begin{figure}[!ht]
  \centering
  \begin{tikzpicture}[
    font=\scriptsize,
    box/.style  = {draw, rounded corners=2pt, align=center,
                   minimum height=5mm, inner sep=2pt},
    sim/.style  = {box, fill=blue!8},
    photo/.style= {draw, rounded corners=2pt, inner sep=0.5pt},
    plabel/.style = {font=\tiny, align=center, inner sep=1pt},
    flow/.style = {-{Stealth[length=1.5mm]}, thick},
    thin flow/.style = {-{Stealth[length=1.3mm]}, semithick, black!60},
  ]

  \node[photo] (q) {\includegraphics[width=0.19\columnwidth]{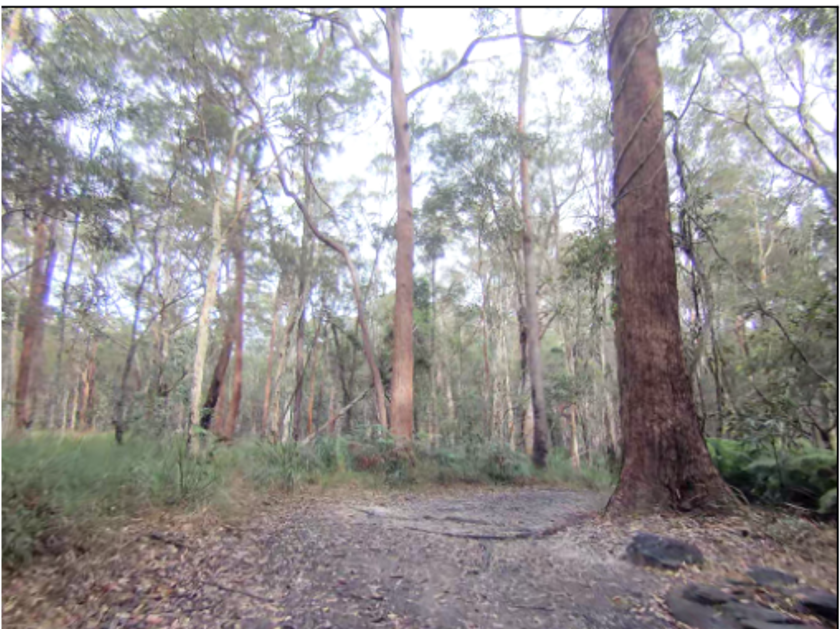}};
  \node[plabel, below=0.8mm of q] {frame $x$};

  \node[photo, right=2.5mm of q, yshift=9mm] (masks)
    {\includegraphics[width=0.20\columnwidth]{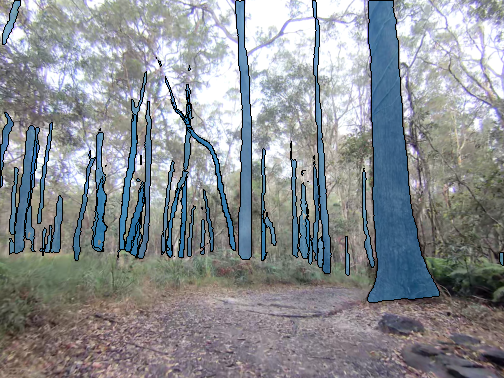}};
  \node[plabel, below=0.8mm of masks] (mc) {SAM3 masks};

  \node[photo, right=2.5mm of q, yshift=-9mm] (depth)
    {\includegraphics[width=0.20\columnwidth]{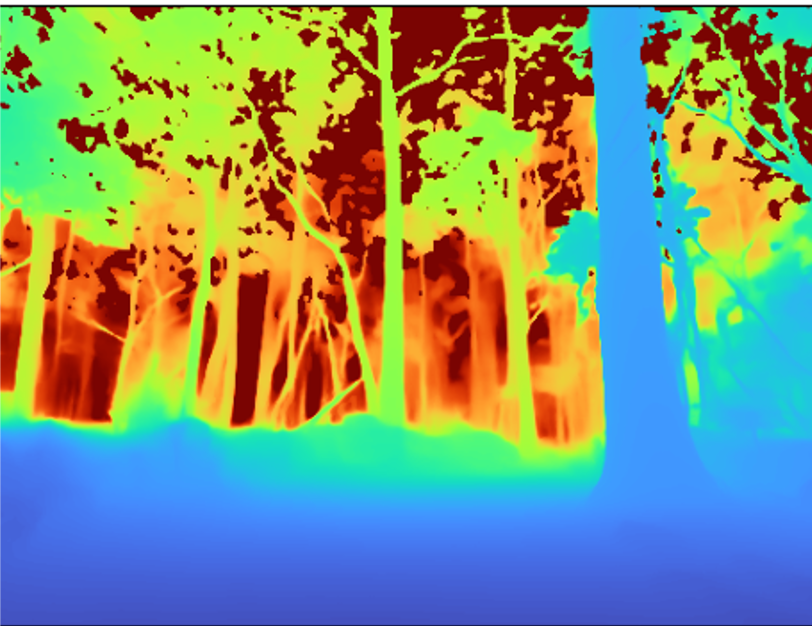}};
  \node[plabel, below=0.8mm of depth] (dc) {DA3 depth};

  \draw[thin flow] (q.east) to[out=40,in=180]  (masks.west);
  \draw[thin flow] (q.east) to[out=-40,in=180] (depth.west);

  \coordinate (mid) at ($(masks.east)!0.5!(depth.east)$);
  \node[box, right=2.5mm of mid, text width=13mm] (fuse)
    {ground-plane projection};

  \draw[thin flow] (masks.east) to[out=0,in=150] (fuse.north west);
  \draw[thin flow] (depth.east) to[out=0,in=210] (fuse.south west);

  \node[photo, right=2.5mm of fuse] (stem)
    {\includegraphics[width=0.25\columnwidth]{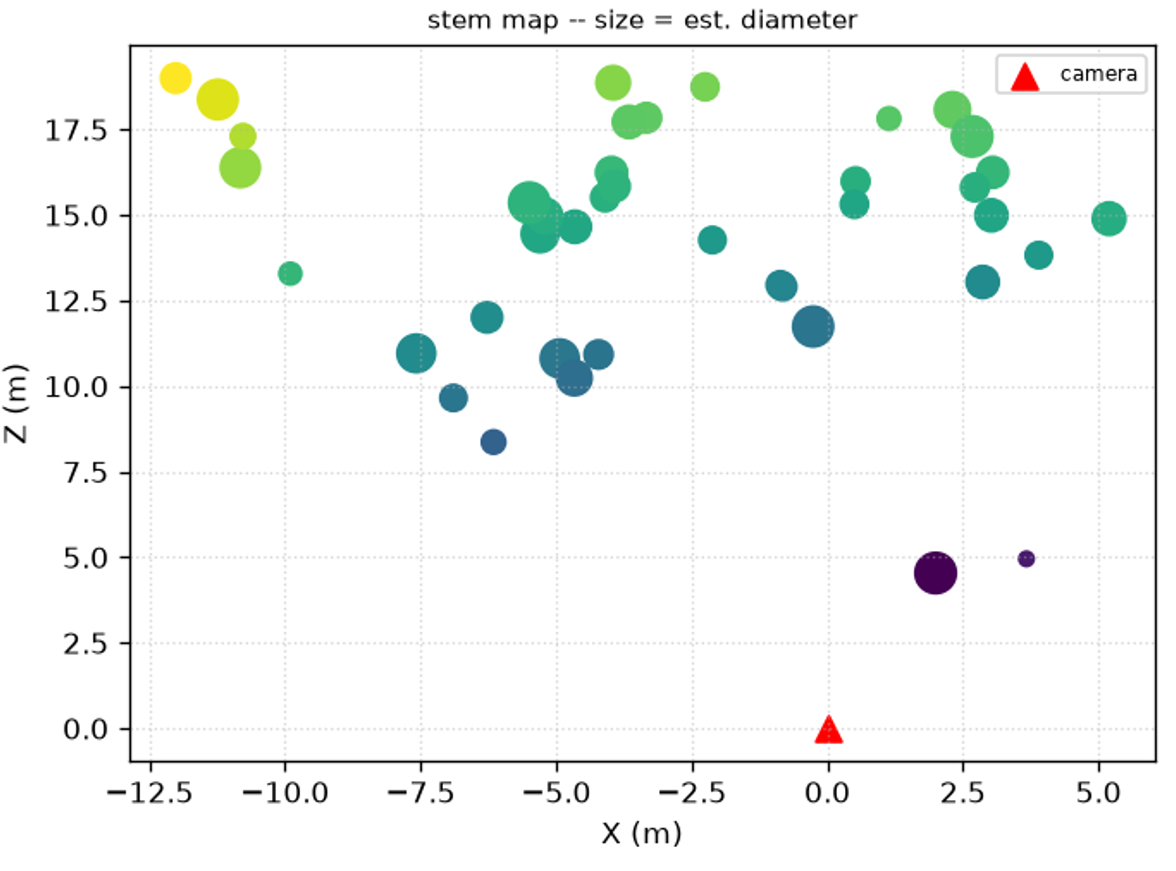}};
  \node[plabel, below=0.8mm of stem] (sc) {stem graph $\mathcal{P}(x)$};

  \draw[flow] (fuse) -- (stem);

  \node[box, below=4mm of sc, xshift=-0.08\columnwidth] (cmp) {ICP / BEV};
  \node[sim, right=2.5mm of cmp] (Sgeom) {$S^{\text{geom}}$};

  \draw[flow] (sc.south) to[out=-90,in=90] (cmp.north);
  \draw[flow] (cmp) -- (Sgeom);

  \end{tikzpicture}
  \caption{\textbf{Stem-graph construction and matching.} SAM3 masks and Depth
    Anything 3 depth are back-projected to place each trunk at a metric
    ground-plane position and diameter, giving the stem graph $\mathcal{P}(x)$. Database frames are encoded the same way, and graphs are compared by ICP or by a blurred bird's-eye-view descriptor to give $S^{\text{geom}}$.}
  \label{fig:stemgraph}
  \vspace{-0.4cm}
  \end{figure}

\fi

\section{Experimental Setup}

\subsection{Dataset}
\label{subsec:dataset}
We evaluate the Venman and Karawatha environments of WildCross, a forest place-recognition benchmark of repeated traversals with SLAM-derived 6DoF poses~\cite{knights2026wildcross}. 
In each environment, the first sequence (V-01, K-01) is the database; V-02/K-02 (reverse traversal), V-03/K-03 (extended route), and V-04/K-04 (repeats $\sim$14 months later) serve as queries. 
From each traversal, we sample $\sim$300 frames by binary search on inter-pose distance for near-uniform spatial coverage. 
Both sequence-matching methods use $L=5$; the slope set $\mathcal{V}$ brackets the median frame-spacing ratio, which ranges from 0.54 (K-04) to 1.64 (V-03).

\subsection{Baselines}
\label{subsec:baselines}

We evaluate three VPR descriptors with complementary design choices:  CosPlace~\cite{berton2022rethinking}, AnyLoc~\cite{keetha2023anyloc}, and MegaLoc~\cite{berton2025megaloc}. This comparison establishes the relative difficulty of the forest traversals across different descriptor families and provides the basis for selecting the appearance baseline used in subsequent experiments. For reranking, we compare against a training-free geometric verification baseline using SuperPoint~\cite{detone2018superpoint} and LightGlue~\cite{lindenberger2023lightglue}, with candidate pairs ranked by the MAGSAC~\cite{barath2020magsac++} inlier count. We apply the verifier to the same candidate sets provided to each VLM reranker.

\subsection{Metrics and Statistical Protocol}
\label{subsec:metrics}
Following WildCross~\cite{knights2026wildcross}, a retrieved image is a true match if its pose lies within 25 m of the query. 
We report on Recall@1 and recall at 100\% precision (R@100P).

\section{Results and Discussion}
\label{sec:results}

\begin{table}[!b]
\centering
{\small\setlength{\tabcolsep}{5pt}
\begin{tabular}{lcccc}
\toprule
(all values in \%) & \multicolumn{2}{c}{Venman} & \multicolumn{2}{c}{Karawatha} \\
\cmidrule(lr){2-3}\cmidrule(lr){4-5}
Single-frame descriptor & R@1 & R@100P & R@1 & R@100P \\
\midrule
CosPlace & 58.7 & 12.3 & 40.5 & 1.8 \\
AnyLoc   & 41.7 & 2.1  & 27.6 & 0.5 \\
MegaLoc  & \textbf{79.0} & \textbf{40.7} & \textbf{59.7} & \textbf{7.3} \\
\bottomrule
\end{tabular}}
\caption{
Single-frame descriptor comparison between CosPlace, AnyLoc, and MegaLoc, averaged over the same-direction query traversals (V-03/V-04, K-03/K-04). Reverse traversals (V-02, K-02) are excluded, as all descriptors perform poorly on them. MegaLoc leads on both metrics in both environments and is used as the base descriptor throughout.
}
\label{tab:singleframe}
\vspace{-0.2cm}
\end{table}

\begin{table*}[!ht]
\centering
{\small\setlength{\tabcolsep}{3pt}
\begin{tabular}{l cc cc cc @{\hspace{10pt}} cc cc cc}
\toprule
& \multicolumn{6}{c}{\textbf{Venman} \quad(database V-01)} & \multicolumn{6}{c}{\textbf{Karawatha} \quad(database K-01)} \\
\cmidrule(lr){2-7}\cmidrule(lr){8-13}
& \multicolumn{2}{c}{V-02} & \multicolumn{2}{c}{V-03} & \multicolumn{2}{c}{V-04}
& \multicolumn{2}{c}{K-02} & \multicolumn{2}{c}{K-03} & \multicolumn{2}{c}{K-04} \\
& \multicolumn{2}{c}{\footnotesize reverse} & \multicolumn{2}{c}{\footnotesize extended} & \multicolumn{2}{c}{\footnotesize $+$14\,mo}
& \multicolumn{2}{c}{\footnotesize reverse} & \multicolumn{2}{c}{\footnotesize extended} & \multicolumn{2}{c}{\footnotesize $+$14\,mo} \\
\cmidrule(lr){2-3}\cmidrule(lr){4-5}\cmidrule(lr){6-7}\cmidrule(lr){8-9}\cmidrule(lr){10-11}\cmidrule(lr){12-13}
 \footnotesize (all values in \%) & R@1 & R@100P & R@1 & R@100P & R@1 & R@100P & R@1 & R@100P & R@1 & R@100P & R@1 & R@100P \\
\midrule
MegaLoc, single frame
 & 38.0 & 19.5 & 70.4 & 30.5 & 87.5 & 50.9
 & 15.4 & 0.4 & 46.6 & 4.5 & 72.9 & 10.2 \\
\addlinespace[2pt]
\midrule
\multicolumn{13}{l}{\textit{MegaLoc $+$ sequence matching (ours)}} \\
\addlinespace[1pt]
\quad Descriptor space
 & \textbf{45.5} & 14.5 & 68.1 & 19.9 & 91.5 & 36.5
 & \textbf{26.2} & 0.4 & 44.8 & 0.0 & 76.3 & 6.4 \\
 \quad Score space
 & 43.8 & \textbf{24.9} & \textbf{75.7} & \textbf{55.3} & \textbf{94.1} & \textbf{71.6}
 & 19.9 & 0.4 & \textbf{71.0} & \textbf{27.1} & \textbf{87.8} & \textbf{62.4} \\
\bottomrule
\end{tabular}}
\caption{
Sequence matching on MegaLoc descriptors ($L = 5$). Score-space matching gives the best R@100P in five of six pairings (all methods tie at the noise floor on K-02) and the best R@1 in four. Descriptor-space matching improves R@1 in four pairings but never improves R@100P over the single-frame baseline. Best per column in bold.
}
\label{tab:seq}
\vspace{-0.2cm}
\end{table*}

\subsection{Baseline Comparison}
\label{subsec:results-baseline}

\begin{figure}
    \centering
    \vspace{-0.2cm}
    \includegraphics[width=\linewidth, trim={0cm 0.2cm 0cm 1.7cm}, clip]{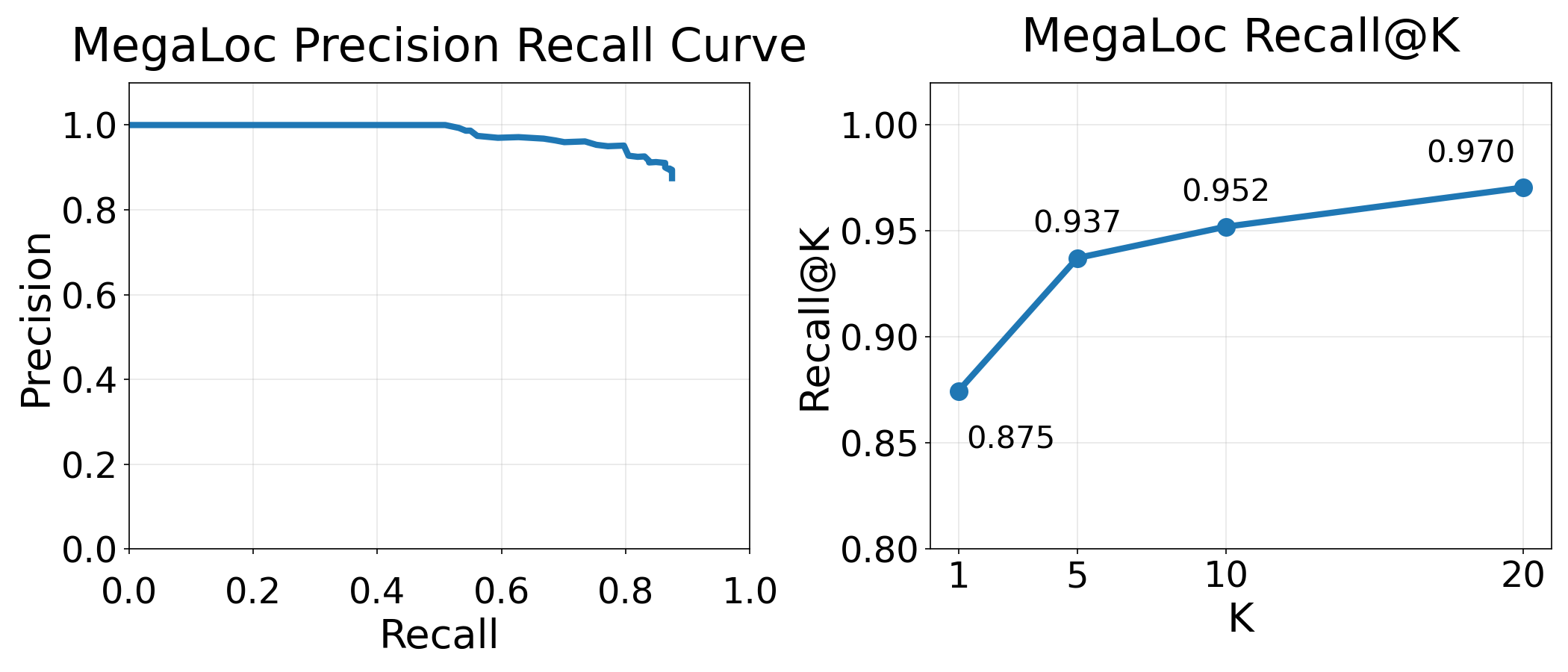}
    \caption{Precision-recall and R@K plots for MegaLoc on the V-01/V-04 dataset (values as fractions in $[0,1]$). %
    }
    \label{fig:results_megaloc_v2}
    \vspace{-0.3cm}
\end{figure}

To select the VPR baseline for the rest of the experiments, we compare three single-frame VPR models on the same-direction query traversals (V-03/V-04, K-03/K-04), averaged per environment (\Cref{tab:singleframe}).
MegaLoc outperforms CosPlace and AnyLoc on both metrics in both environments (R@1 79.0\% vs. 58.7\% and 41.7\% on Venman; 59.7\% vs. 40.5\% and 27.6\% on Karawatha; best R@100P 40.7\% and 7.3\%), so we use it as the base descriptor throughout. AnyLoc performs the worst despite targeting unstructured environments. Reverse traversals are poor for all descriptors. On V-04, 34 of 271 MegaLoc queries fail at top-1 (\Cref{fig:results_megaloc_v2}), mostly through visual self-similarity, where repeated trail framing makes a wrong frame the closest match, or through glare and viewpoint change pushing the correct frame away in descriptor space.

\subsection{Temporal Stream: Sequence Matching}
\label{subsec:results-seq}

\label{subsec:results-seq-compare}
The score-space sequence matching matched or exceeded single-frame R@1 and R@100P in all six pairings (\Cref{tab:seq}). Descriptor-space matching improved R@1 in four of the six pairings but did not improve on the single-frame R@100P in any. Its R@1 loss is confined to the two extended-route pairings (V-03, K-03). This suggests that a fixed frame-count window cannot adapt when one traversal covers more (or less) distance per frame than the one it is matched against. On V-03, for example, each query frame spans 1.64 reference frames.

The descriptor-space precision drop follows from the temporal averaging it applies. Descriptors are drawn towards the local mean of their sequence window, so distinct but visually similar places converge, and the margin between true and false matches narrows. R@1 requires only the correct match to score highest for each query, whereas R@100P requires a single threshold that separates correct from incorrect matches across all queries, so a narrower margin lowers R@100P more than it lowers R@1. On V-03, this convergence is severe enough to displace the correct match from the top rank. Score-space matching avoids this as it never changes the descriptors, using temporal evidence only to confirm a match.

For loop closure, where precision matters most, score-space matching is the more suitable of the two, achieving the best R@100P in five of six pairings (tying on K-02) while also leading R@1 in four. These methods can reach similar R@1 yet differ sharply in R@100P. On V-04, R@1 is \percentage{.941} vs \percentage{.915}, but R@100P is \percentage{.716} vs \percentage{.365}. Hence, sequence matching front ends for loop closure should be selected based on precision-aware metrics instead of just recall.

\subsubsection{Sequence Length Ablation}
\label{subsec:results-seq-length}

\begin{figure}
    \centering
    \vspace{-0.2cm}
    \includegraphics[width=0.8\linewidth, trim={0cm 0.2cm 0cm 3.8cm}, clip]{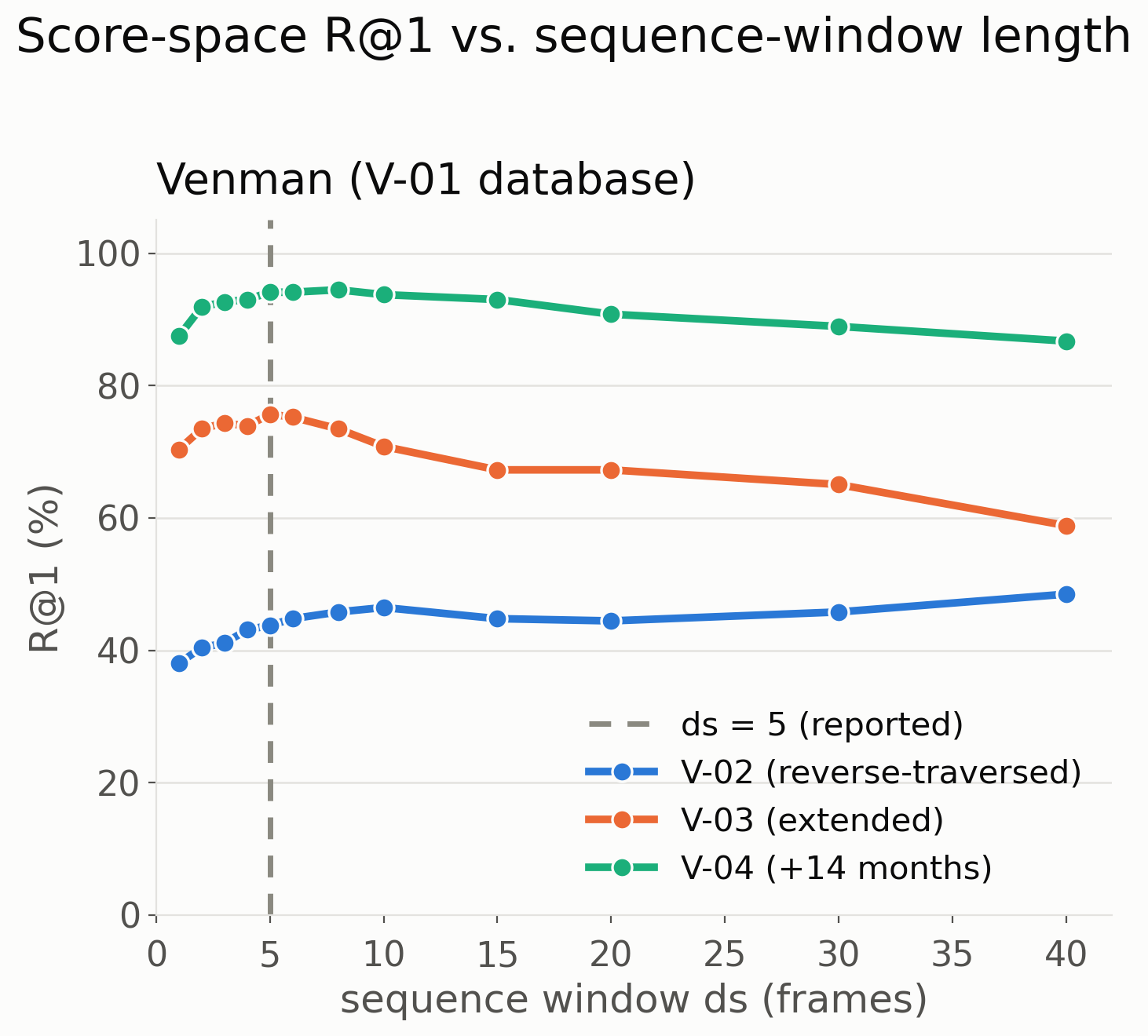}
    \caption{Score-space R@1 on Venman (database V-01) across varying sequence lengths $L$, with same settings as in \Cref{tab:seq}. $L = 1$ corresponds to the single-frame baseline; the dashed line, $L = 5$, is chosen for all other experiments.}
    \label{fig:seq_length_ablation}
    \vspace{-0.3cm}
\end{figure}

To examine the impact of the sequence length on the score-space performance, we vary $L$ from 1 to 40 on Venman (\Cref{fig:seq_length_ablation}). All pairings improve up to around $L = 5$; beyond this, V-03 and V-04 degrade, while V-02 continues to improve slowly. $L = 5$ therefore performs best or near-best in all three pairings.
Most of the gain arrives within the first few frames, whereas longer windows increasingly span points where the query route leaves or rejoins the reference route, breaking the straight-line trajectory assumed in \cref{eq:seqslam}. V-03 and V-04 each leave the reference route for part of the traverse, whereas V-02 covers the same reference route, mostly in reverse, so none of its windows span such a point.

\subsection{Complementary Representations}
\label{subsec:results-complementary}

\subsubsection{Linguistic Stream: Captions and Reranking}
\label{subsec:results-vlm}

\begin{figure}[!ht]
    \centering
    \vspace{-0.2cm}
    \includegraphics[width=\linewidth,  trim={0cm 0.5cm 0cm 2.5cm}, clip]{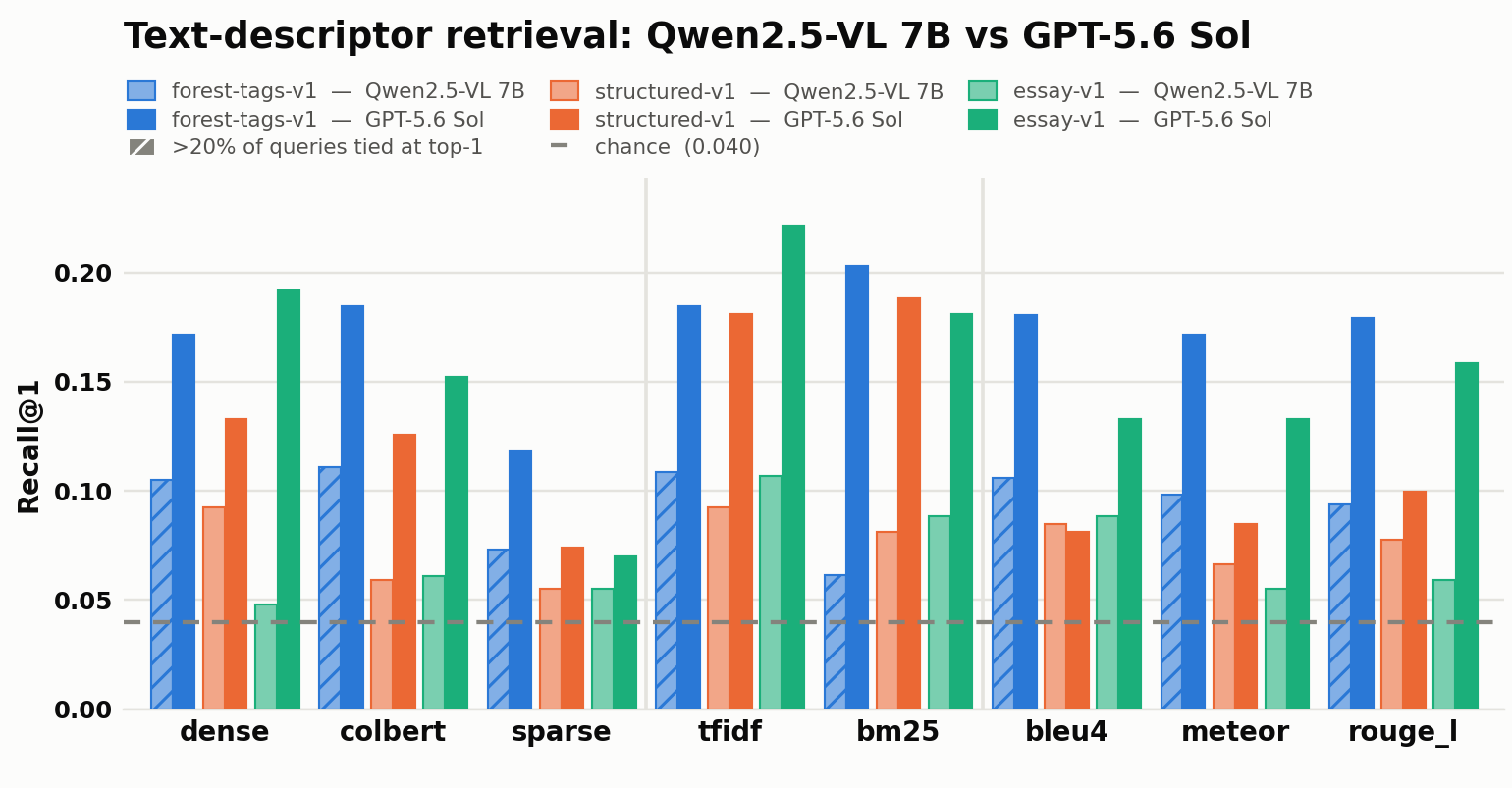}
    \caption{
    Caption-based retrieval R@1 on V-04 queries against V-01 (271 valid queries) for eight scoring functions $\phi$ (x-axis), three prompt schemas $\pi$ (colour: \textcolor{RoyalBlue}{forest tags}, \textcolor{orange}{structured}, {\color{ForestGreen}essay}) and two caption models $\Gamma$ (shade: lighter for Qwen2.5-VL 7B, darker for GPT-5.6 Sol). The best cell (\percentage{0.221}) is far below MegaLoc (\percentage{0.875}); chance is \percentage{0.040}.
    }
    \label{fig:results_text_descriptors}
    \vspace{-0.3cm}
\end{figure}

\paragraph{Caption Retrieval.}
Caption-based retrieval stayed far below the image baseline in our setup, and not because of the scorer. The best cell in the full sweep (\Cref{fig:results_text_descriptors}) is TF-IDF over GPT-5.6 Sol~\cite{openai2026gpt56} essay captions at R@1 $=$ \percentage{0.221} (chance $=$ \percentage{0.040}, MegaLoc $=$ \percentage{0.875}), and R@100P is 0.0\% in all eighteen runs scored with the captioning metrics. Every scoring family lands in the same narrow band despite scoring by quite different rules. A repeatability control locates the cause upstream. By BLEU-4, METEOR and ROUGE-L, captions of the same place are only $1.00-1.09\times$ as similar to each other as captions of different places.
The captions carry little place-specific information; consequently, none of these scoring rules can reliably separate places.
This quantifies, in a forest setting, the argument of~\cite{cheng2025scale} that an intermediate description discards the cues that separate similar places. For text retrieval to become useful, captions would need to consistently name persistent, place-specific attributes.

\paragraph{VLM Reranking.}

Reranking is applied only to gated queries (\Cref{tab:rerank}). 
The best configuration, Qwen3-VL 32B with the failure gate and tie band, fixes 10 baseline errors and breaks 3, raising R@1 from \percentage{0.875} to \percentage{0.900} ($p = 0.092$) but lowering R@100P from \percentage{0.509} to \percentage{0.399}. GPT-5.6 Sol is weaker (R@1 $=$ \percentage{0.886}), and widening its shortlist to a fixed top-10 only changes R@1 to \percentage{0.889}. SuperPoint+LightGlue reach the same R@1 as Qwen3-VL 32B with a higher R@100P (\percentage{0.535}), yet the two agree on the top-1 for only 20 of the 52 queries. The VLM signal is partly complementary to geometric verification but, on this data, not better than it.

\begin{table}[!ht]
\centering
{\small\setlength{\tabcolsep}{3pt}
\begin{tabular}{lcccccc}
\toprule
Run & Q & Fix/ & R@1 & R@100P & Ceil. & $p$ \\
(all values in \%)    & rr & brk &     &        & R@1 & \\
\midrule
MegaLoc baseline & -- & -- & 87.5 & 50.9 & -- & -- \\
\quad SP+LightGlue & 52 & 9/2 & \textbf{90.0} & \textbf{53.5} & 93.0 & 6.5 \\
\midrule
\multicolumn{7}{l}{Gated, tie-band} \\
\quad Qwen3-VL 32B & 52 & 10/3 & \textbf{90.0} & 39.9 & 93.0 & 9.2 \\
\quad GPT-5.6 Sol & 52 & 7/4 & 88.6 & 50.9 & 93.0 & 54.9 \\
\multicolumn{7}{l}{Margin gate} \\
\quad Qwen3-VL 32B & 130 & 10/4 & 89.7 & 39.9 & 93.0 & 18.0 \\
\multicolumn{7}{l}{Gated, top-10} \\
\quad GPT-5.6 Sol & 58 & 8/4 & 88.9 & 50.9 & \textbf{95.2} & 38.8 \\
\bottomrule
\end{tabular}}
\caption{
Reranking MegaLoc (V-01 reference, V-04 queries, 271 queries), grouped by gate and candidate picker. $Q$: queries reranked; Fix/Brk: errors fixed / correct matches broken; Ceil.: R@1 of a perfect reranker; $p$: McNemar test on discordant queries against the baseline VPR. None is significant at $\alpha = 0.05$.
}
\label{tab:rerank}
\vspace{-0.4cm}
\end{table}

\paragraph{Reranking Failure Gate.}

Of the 34 baseline misses, the gate passes 30, the tie-band shortlist retains a correct frame for 15 of those, and the VLM converts 10. Four misses are lost to the gate, fifteen to the candidate picker and five to the reranker. The gate reaches AUC-PR \percentage{0.879}, against \percentage{0.374} for the similarity-margin heuristic. %
It catches the same 30 misses as the margin gate while passing 22 rather than 100 already-correct queries, cutting VLM calls by 60\% with no loss in R@1 (\percentage{0.900} vs. \percentage{0.897}). The candidate picker can only select by position in the ranking that already failed, and widening it to a fixed top-10 raises the ceiling only from \percentage{0.930} to \percentage{0.952}. Raising this ceiling needs candidate selection that draws on information the baseline ranking lacks, such as temporal context or a second descriptor.

\begin{figure}[t]
  \centering
  \includegraphics[width=0.68\columnwidth]{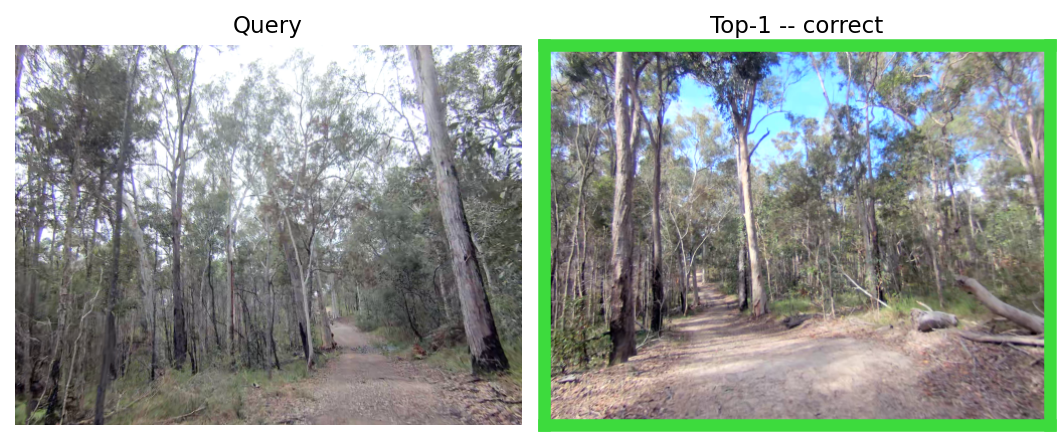}\\[2pt]
  \includegraphics[width=\columnwidth]{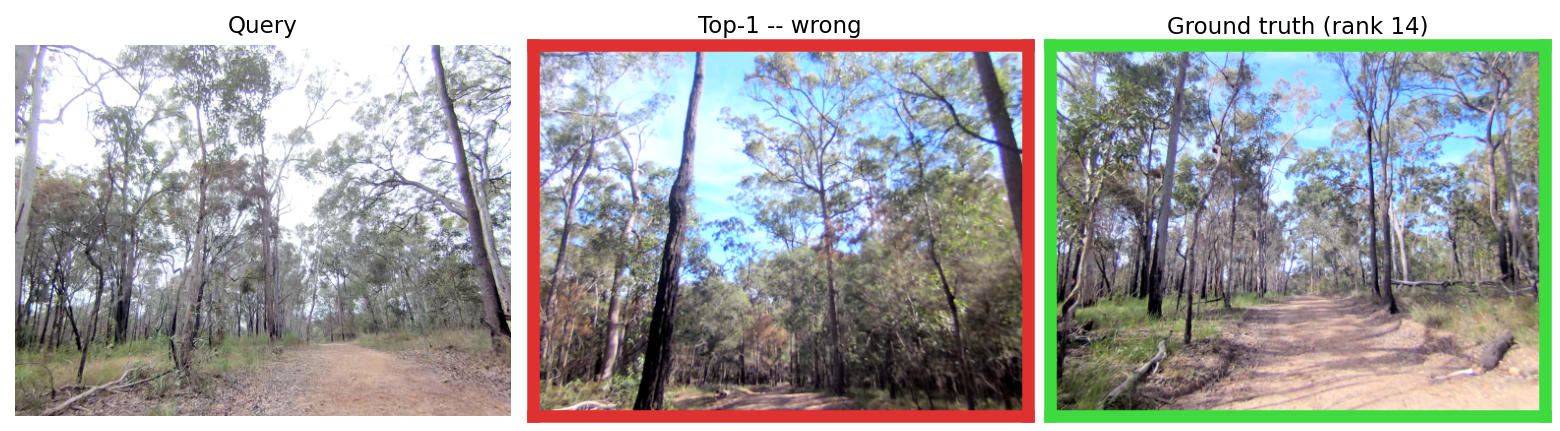}
  \caption{Qualitative examples with the class-binned scene-graph
  descriptor (Venman, V-04 queries against V-01). Top: a correct top-1
  match. Bottom: a failure, where the top-1 candidate is a different place
  with similar trail framing and the nearest ground-truth reference is ranked 14th.
  }
  \label{fig:qualitative}
  \vspace{-0.4cm}
\end{figure}

\ifSG
\subsubsection{Structural Stream: Scene Graphs} %
\label{subsec:results-sg}

\paragraph{Scene-Graph Descriptors.}
None of the scene-graph descriptors approaches the MegaLoc baseline on any Venman pairing (\Cref{tab:struct}). The two appearance-based descriptors, mean-pool and class-binned, consistently outperform layout, so most of the signal comes from object appearance rather than 2D arrangement. Fusing all three descriptors with equal weights, $S^{\text{struct}}$, gives no consistent gain, adding at most \percentage{0.030} R@1 over the best single descriptor and falling below mean-pool on V-03. However, on the reverse traversal $S^{\text{struct}}$ degrades less than MegaLoc, dropping from \percentage{0.240} to \percentage{0.212} against \percentage{0.875} to \percentage{0.380} relative to V-04. This suggests that it may be useful as a complement to the baseline on reverse traversals, which we leave for future work.
Figure~\ref{fig:qualitative} shows a class-binned failure. The top-1 candidate shares the query's trail framing but lies at a different place, while the true reference is ranked 14th.

\begin{table}[!ht]
\centering
{\small\setlength{\tabcolsep}{6pt}
\begin{tabular}{lccc}
\toprule
Descriptor & V-02 & V-03 & V-04 \\
\midrule
Chance       & 4.1 & 3.5 & 4.0 \\
MegaLoc      & 38.0 & 70.4 & 87.5 \\
\midrule
Layout       &  10.9 & 9.3 & 11.6 \\
Mean-pool    & 16.2 & \textbf{18.5} & 19.9 \\
Class-binned & 18.2 & 16.2 & 22.6 \\
\midrule
Fused        & \textbf{21.2} & 17.5 & \textbf{24.0} \\
\bottomrule
\end{tabular}}
\caption{Structural stream R@1 (\%) on Venman (database V-01). All descriptors are read from the same scene graphs with DINOv2 features and fused with equal weights.}
\label{tab:struct}
\vspace{-0.4cm}
\end{table}

\paragraph{Segment Reranking.}
Reordering the baseline's top ten candidates by warped mask IoU (\Cref{subsec:sg-method}) lowers R@1 in all three Venman pairings (\Cref{tab:stemgraph}), from \percentage{0.875} to \percentage{0.694} on V-04. No valid homography is found for a third of candidate pairs, which are left unscored and ranked last. Where one is found, mask overlap separates true from false matches (mean IoU \percentage{0.327} vs. \percentage{0.212}), but more weakly than the appearance similarity it replaces. A single homography cannot align non-planar forest scenes, which motivates comparing trunk positions in metric ground-plane coordinates.

\subsubsection{Geometric Stream: Stem Graphs}
\label{subsec:results-geom}

\paragraph{Stem Graph Matching.}
Using no appearance features, stem-graph retrieval on V-04 reaches R@1 of \percentage{0.151} with the BEV descriptor ($S^{\text{bev}}$), $3.8\times$ chance, while ICP ($S^{\text{icp}}$) reaches \percentage{0.059}, close to chance (\percentage{0.040}). Both remained far below the baseline in all three pairings (\Cref{tab:stemgraph}). Tightening the range cut from 25 m to 15 m raises the number of reference frames with no trunks, which can never be retrieved, from 10 to 26 of 302, so we keep the 25 m cut even though distant trunks have the least reliable depth.

\paragraph{Stem Graph Construction.}%
Transforming each query graph into its ground-truth match using the recorded poses gives a median trunk position error of 2.09 m, and 2.03 m even for frame pairs under 5 m apart, so the error comes mainly from per-frame reconstruction instead of viewpoint change. 
At this error, the nearest database trunk can be the wrong tree, which can mislead ICP. However, rebuilding the V-04 stem graphs from the benchmark's LiDAR depth raised ICP only to \percentage{0.078} and lowered BEV to \percentage{0.059} (\Cref{tab:stemgraph}), so monocular depth error alone does not explain the low performance. The BEV drop may partly reflect its blur width, which was set for monocular depth error. Other likely limits are the level-camera assumption, missed or merged trunk masks, and trunk layouts that repeat across places.

\begin{table}[!ht]
\centering
{\small\setlength{\tabcolsep}{6pt}
\begin{tabular}{lccc}
\toprule
Method & V-02 & V-03 & V-04 \\
\midrule
Chance                   & 4.1 & 3.5 & 4.0 \\
Homography re-rank       & 31.0 & 56.2 & 69.4 \\
\midrule
Stem-graph ICP           & 5.7 & 6.2 & 5.9 \\
Stem-graph BEV           & \textbf{9.8} & 7.1 & \textbf{15.1} \\
\midrule
Stem-graph ICP (LiDAR)   & 5.7 & 4.0 & 7.8 \\
Stem-graph BEV (LiDAR)   & 9.1 & \textbf{9.3} & 5.9 \\
\bottomrule
\end{tabular}}
\caption{R@1 (\%) of segment reranking and stem-graph retrieval on Venman (database V-01). The homography rerank reorders MegaLoc's top-10 candidates by warped mask IoU. The stem-graph rows retrieve over the full database from only trunk positions. All fall below the baseline across all pairings. 
LiDAR rows rebuild the stem graphs from the benchmark's LiDAR depth instead of monocular depth.
}
\vspace{-0.4cm}
\label{tab:stemgraph}
\end{table}

\fi

\section{Conclusion and Future Directions}
\label{sec:conclusion}

We studied whether temporal, linguistic, structural or geometric context could improve forest VPR  on the WildCross dataset. Score-space sequence matching matched or exceeded the MegaLoc baseline on both metrics across all pairings. While descriptor-space matching reached similar recall, it was less precise. In our implementation, captions barely distinguished places, and reranking was bounded by the shortlist rather than the gate\ifSG{}; 2D layout added little over appearance, and trunk layout matching remained weak even with LiDAR depth. Overall, temporal order was the only added information that clearly improved on the baseline in our experiments, whereas captions\ifSG{} and 2D layout\fi{}, which re-express appearance in a more compressed form, gave little.

Temporal context was most effective when individual frames retained sufficient appearance similarity. On extended routes, sequence matching compensated for differences in traversal rate, substantially improving high-precision retrieval. Reverse traversal remained considerably more challenging because opposite-facing observations often shared insufficient visual content for temporal aggregation to recover reliable correspondences. This suggests that stronger viewpoint-invariant representations are required before temporal consistency can be fully exploited under severe viewpoint change. The complementary representations were evaluated zero-shot and, for the more computationally expensive streams, on the Venman subset of WildCross. Segment reranking and stem-graph retrieval were additionally evaluated on the reverse and extended traversals, where both remained below the appearance baseline. The observed VLM reranking improvement was not statistically significant at the current number of queries. Moreover, the traversals were subsampled for computational tractability; evaluation against a denser database and under tighter ground truth tolerance might be more challenging. 
Finally, both environments are eucalypt forests in a single climatic region. The generalisation of our findings to forests with different vegetation densities, terrain structures or seasonal characteristics is left for future evaluation on independent datasets.
Future work should investigate language as a complementary similarity signal instead of replacing visual similarity~\cite{idan2026lavpr}, incorporate depth into structural representations, and explore geometry-aware refinement of appearance descriptors~\cite{nedov2026geometry}.

\paragraph{Acknowledgements.}
This research was partially supported by ARC Laureate Fellowship FL210100156 to MM. MM and SH acknowledge continued support from the QUT Centre for Robotics, and Monash and CSIRO authors acknowledge support from their institutions. 
This work was supported by Monash eResearch capabilities, including M3. The authors used Claude (Anthropic) to assist with writing, coding, and figures and tables, and reviewed and revised all AI-assisted content, taking full responsibility for the paper.
\bibliographystyle{named-etal}
\bibliography{references}

\end{document}